\pdfoutput=1

\documentclass[11pt]{article}

\usepackage[final]{acl}

\usepackage{times}
\usepackage{latexsym}

\usepackage[T1]{fontenc}

\usepackage[utf8]{inputenc}

\usepackage{microtype}

\usepackage{inconsolata}

\usepackage[utf8]{inputenc} 
\usepackage[T1]{fontenc}    
\usepackage{hyperref}       
\usepackage{url}            
\usepackage{booktabs}       
\usepackage{amsfonts}       
\usepackage{nicefrac}       
\usepackage{xcolor}         
\usepackage{xspace}
\usepackage{graphicx}
\usepackage{amsmath}    
\usepackage{bm}
\usepackage{arydshln}
\usepackage{subfigure}
\usepackage{enumitem}
\usepackage{multirow}
\usepackage{algorithm2e}
\usepackage{setspace}
\usepackage{amssymb}
\usepackage{caption}
\usepackage{cleveref}
\usepackage{colortbl}
\usepackage{listings}
\usepackage{color}

\usepackage[most]{tcolorbox}

\newtcolorbox{genbox}[1][]{
  colback=white,
  colframe=black,
  sharp corners,
  boxrule=1pt,
  left=10pt,
  right=10pt,
  top=10pt,
  bottom=10pt,
  #1
}

\newtcolorbox{mybox}[2][]{
  colback=white,
  colframe=black,
  fonttitle=\bfseries,
  coltitle=black,
  colbacktitle=white,
  enhanced,
  attach boxed title to top left={yshift=-2mm, xshift=1cm},
  boxed title style={colframe=white, colback=white},
  title=#2,
  #1
}

\newcommand{\data}{\textsc{GRBench}\xspace}
\newcommand{\Ours}{\textsc{Graph-CoT}\xspace}

\definecolor{codegreen}{rgb}{0,0.6,0}
\definecolor{codegray}{rgb}{0.5,0.5,0.5}
\definecolor{codepurple}{rgb}{0.58,0,0.82}
\definecolor{backcolour}{rgb}{0.95,0.95,0.92}

\lstdefinestyle{mystyle}{
    backgroundcolor=\color{backcolour},   
    commentstyle=\color{codegreen},
    keywordstyle=\color{magenta},
    numberstyle=\tiny\color{codegray},
    stringstyle=\color{codepurple},
    basicstyle=\ttfamily\footnotesize,
    breakatwhitespace=false,         
    breaklines=true,                 
    captionpos=b,                    
    keepspaces=true,                 
    numbers=left,                    
    numbersep=5pt,                  
    showspaces=false,                
    showstringspaces=false,
    showtabs=false,                  
    tabsize=2
}

\lstdefinelanguage{json}{
    basicstyle=\normalfont\ttfamily,
    numbers=left,
    numberstyle=\scriptsize,
    stepnumber=1,
    numbersep=8pt,
    showstringspaces=false,
    breaklines=true,
    frame=lines,
    backgroundcolor=\color{white},
    literate=
     *{0}{{{\color{blue}0}}}{1}
      {1}{{{\color{blue}1}}}{1}
      {2}{{{\color{blue}2}}}{1}
      {3}{{{\color{blue}3}}}{1}
      {4}{{{\color{blue}4}}}{1}
      {5}{{{\color{blue}5}}}{1}
      {6}{{{\color{blue}6}}}{1}
      {7}{{{\color{blue}7}}}{1}
      {8}{{{\color{blue}8}}}{1}
      {9}{{{\color{blue}9}}}{1}
      {:}{{{\color{purple}{:}}}}{1}
      {,}{{{\color{purple}{,}}}}{1}
      {\{}{{{\color{brown}{\{}}}}{1}
      {\}}{{{\color{brown}{\}}}}}{1}
      {[}{{{\color{brown}{[}}}}{1}
      {]}{{{\color{brown}{]}}}}{1},
}

\title{Graph Chain-of-Thought: Augmenting Large Language Models by Reasoning on Graphs}

\author{Bowen Jin$^1$, Chulin Xie$^1$, Jiawei Zhang$^1$, Kashob Kumar Roy$^1$, Yu Zhang$^1$ \\ \textbf{Zheng Li$^2$, Ruirui Li$^2$, Xianfeng Tang$^2$, Suhang Wang$^3$, Yu Meng$^4$, Jiawei Han$^1$} \\
  $^1$University of Illinois at Urbana-Champaign, $^2$Amazon \\
  $^3$Pennsylvania State University, 
  $^4$University of Virginia \\
  \texttt{\small bowenj4@illinois.edu}}

\begin{document}

\maketitle

\begin{abstract}
Large language models (LLMs), while exhibiting exceptional performance, suffer from hallucinations, especially on knowledge-intensive tasks.
Existing works propose to augment LLMs with individual text units retrieved from external knowledge corpora to alleviate the issue.
However, in many domains, texts are interconnected (\textit{e.g.}, academic papers in a bibliographic graph are linked by citations and co-authorships) which form a (text-attributed) graph. The knowledge in such graphs is encoded not only in single texts/nodes but also in their associated connections.
To facilitate the research of augmenting LLMs with graphs, we manually construct a \textbf{G}raph \textbf{R}easoning Benchmark dataset called \data, containing 1,740 questions that can be answered with the knowledge from 10 domain graphs.
Then, we propose a simple and effective framework called Graph Chain-of-thought (\Ours) to augment LLMs with graphs by encouraging LLMs to reason on the graph iteratively. Each \Ours iteration consists of three sub-steps: LLM reasoning, LLM-graph interaction, and graph execution.
We conduct systematic experiments with three LLM backbones on \data, where \Ours outperforms the baselines consistently. 
The code is available at \url{https://github.com/PeterGriffinJin/Graph-CoT}.
\end{abstract}

\section{Introduction}

\begin{figure}[t]
\centering
\includegraphics[width=0.49\textwidth]{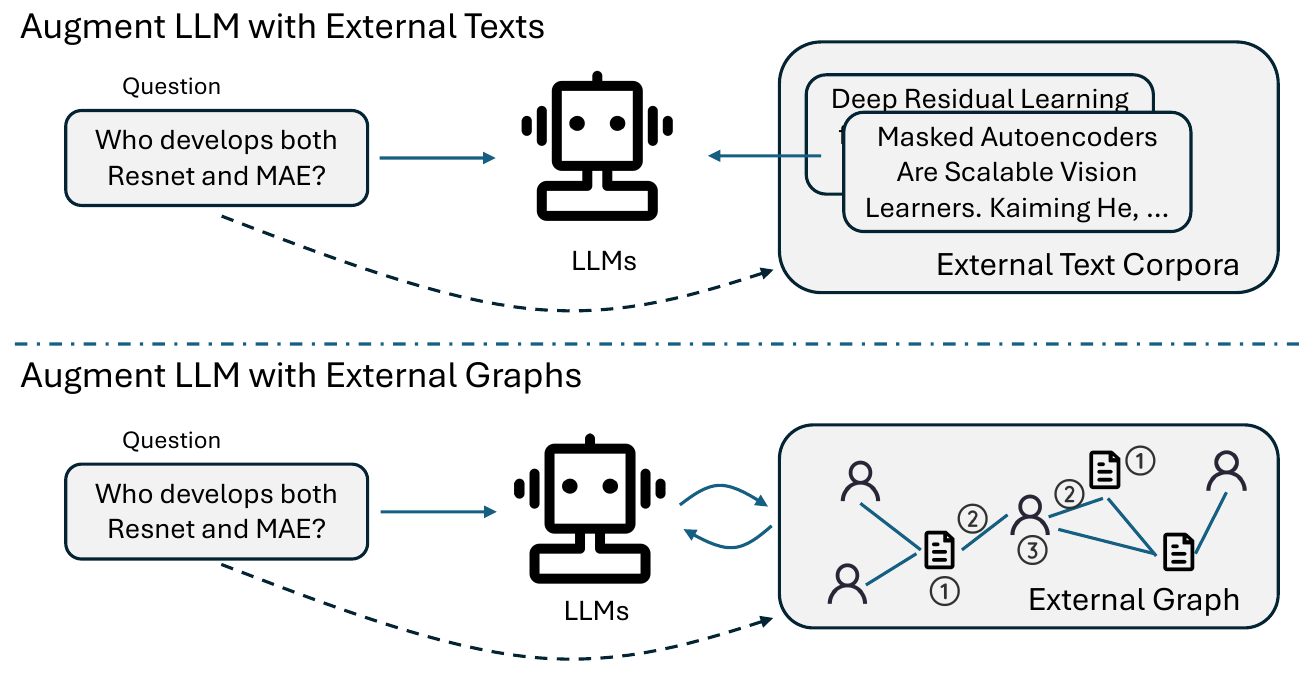}
\caption{Augmenting LLMs with external text corpora or external text-attributed graph.}\label{intro-figure}
\vspace{-0.25in}
\end{figure}

Large language models (LLMs) \cite{touvron2023llama,albert2024mixtral} have demonstrated their exceptional language understanding and text generation capability in real-world scenarios \cite{zhao2023survey}.
However, LLMs suffer from hallucination problems and sometimes tend to generate content that appears plausible but is ungrounded \cite{tonmoy2024comprehensive}.
This is because they memorize world knowledge parametrically and fail to refer to concrete knowledge sources \cite{zhang2023siren}.
To alleviate the hallucination issues, existing works propose to augment LLMs with external text corpora as knowledge sources \cite{shuster2021retrieval,wu2023ragtruth} and treat every single document as a knowledge unit. 
Retrieval augmentation (RAG) \cite{lewis2020retrieval, gao2023retrieval} is then proposed to enable LLMs to interact with external knowledge sources, where relevant texts are retrieved and serve as contexts to improve the factuality of LLMs (shown in Figure \ref{intro-figure} (a)).
However, retrieval augmentation assumes that knowledge is well represented in individual text units and ignores the correlations among multiple text units.

In real-world scenarios, text units are generally interconnected, forming a (text-attributed) graph.
The knowledge of such graphs is reflected not only in the form of texts but also in the structure of their connections.
For example, academic papers in a bibliographic graph are linked by citation links \cite{wang2020microsoft}. We can trace the source of a research direction \cite{bai2019scientific} by traversing such a graph. Cases and opinions in a legal graph are interconnected by reference edges \cite{sadeghian2018automatic}. We can verify the judgment for a case by looking up its citations on such a graph \cite{chen2019learning}.

Although widely used for text corpora as external knowledge sources, retrieval-augmentation cannot be readily used to augment LLMs with graphs for two reasons: 1) \textit{Structure Context}: Retrieval augmentation can find individual nodes/texts from the graphs which can serve as context to augment the LLMs. However, knowledge on the graph also lies in the structure which can not be captured by single nodes/texts. 2) \textit{Graph Size Explosion}: Although it is feasible to convert local subgraph structures into text descriptions as the input contexts to LLMs, the size of the local subgraph increases exponentially as the hop number increases, resulting in an excessively long context sequence. This could cause LLMs to be lost in the middle \cite{liu2023lost} given a plethora of irrelevant information in the context. In addition, the long sequence could potentially exceed the input length limitations of LLMs  \cite{zhao2023survey}.

Therefore, it is an important research topic to augment LLMs with such graph information. Unfortunately, there has been a lack of benchmark datasets to support the development of methodology and facilitate the evaluation of the proposed models. 
To this end, we first construct a \textbf{G}raph \textbf{R}easoning benchmark dataset called \data. 
\data includes ten real-world graphs that can serve as external knowledge sources for LLMs from five domains including academic, e-commerce, literature, healthcare, and legal domains. Each sample in \data consists of a manually designed question and an answer, which can be directly answered by referring to the graphs or retrieving the information from the graphs as context. To make the dataset comprehensive, we include samples of different difficulty levels: easy questions (which can be answered with single-hop reasoning on graphs), medium questions (which necessitate multi-hop reasoning on graphs), and hard questions (which call for inductive reasoning with information on graphs as context).

We propose a simple and effective framework called Graph Chain-of-thought (\Ours). The main idea is to enable LLMs to traverse the graph step-by-step to figure out the key information needed, rather than directly feeding the whole subgraph as context into the LLMs (shown in Figure \ref{intro-figure} (b)). \Ours is an iterative framework, where one iteration corresponds to one step on the graph. Each iteration in \Ours consists of three sub-steps: 1) \textit{Reasoning}: LLMs propose what conclusion we can make with the current information and what further information is needed from the graph; 2) \textit{Interaction}: LLMs generate the interactions needed to fetch information from the graph (\textit{e.g.}, finding the nodes, checking the neighbors, \textit{etc}); 3) \textit{Execution}: The requests from the interaction step are executed on the graph and the corresponding information is returned.
In this way, LLMs can conduct chain-based reasoning on the graph and find the key information on the graph.
This process will be iterated until LLMs conclude the final answer in the reasoning sub-step.

In summary, our contributions are as follows:
\begin{itemize}[leftmargin=*,nosep]
    \item We propose the problem of augmenting LLMs with external graphs and introduce a comprehensive benchmark dataset called \data.
    \item We develop a straightforward and effective framework \Ours to encourage the LLMs to reason on the graph iteratively.
    \item We conduct extensive experiments on \data to demonstrate the effectiveness of \Ours and analyze its performance across different demonstration settings, backbone LLMs, and questions difficulties. Furthermore, we explore its failure cases with future directions outlined.
\end{itemize}






\section{Preliminaries}\label{sec::profdef}

\vspace{3px}
\noindent\textbf{Definition 2.1. Graph.}
A graph can be denoted as $\mathcal{G}=(\mathcal{V}, \mathcal{E})$, where $\mathcal{V}$ and $\mathcal{E}$ are node set and edge set, respectively. Each $v_i\in \mathcal{V}$ can be associated with some feature information $\mathcal{X}_{v_i}$. 
For example, in an e-commerce item graph, $v\in\mathcal{V}$ are items, $e\in\mathcal{E}$ are co-purchase edges, and $\mathcal{X}_v$ include features such as item title, description, price, and category. 
\textit{In this work, we formulate all the features as texts and the graph is also called a text-attributed graph.}


\noindent\textbf{Definition 2.2. Neighbors and Degree.}
The \textit{neighbors} of a node $v_i$ refer to nodes which are linked to $v_i$ on the graph, denoted as $N(v_i)=\{v_j|e_{v_i,v_j}\in \mathcal{E}\}$.
The \textit{degree} of a node $v_i$ refers to the number of $v_i$'s neighbors, denoted as $D(v_i)=|N(v_i)|$.



\section{\data Dataset}


\subsection{Dataset Overview}
We create the \data dataset to evaluate how effectively LLMs can interact with domain-specific graphs containing rich knowledge to solve the desired problem.
\data contains 10 graphs from 5 general domains (academia, e-commerce, literature, healthcare, and legal).
Each data sample in \data is a question-answer pair.
The questions are designed to simulate the real-world use cases in specific domains. However, it is hard for LLMs to answer those questions directly with their internal knowledge stored in model parameters; they need to interact with external domain-specific graphs.
The overall statistics of \data are in Table \ref{tb-data}.

\begin{table}
\centering 
\caption{Dataset Statistics of \data.}\label{tb-data}
\vspace{-0.15in}
\renewcommand\arraystretch{0.88}
\fontsize{8}{10}\selectfont \setlength{\tabcolsep}{0.4em}
\scalebox{0.8}{
\begin{tabular}{@{}cccc cc@{}}
\toprule
\multirow{2}{*}{Domain}   & \multirow{2}{*}{Topic} & \multicolumn{2}{c}{Graph Statistics} & \multicolumn{2}{c}{Data}      \\ 
\cmidrule(lr){3-4} \cmidrule(lr){5-6} 
                          &                        & \# Nodes      & \# Edges         & \# Templates & \# Questions  \\ \midrule
\multirow{6}{*}{Academic} & CS                 &  \textasciitilde 8M    & \textasciitilde 52M      & 15           & 150               \\
\cmidrule(lr){2-6}
                          & Biology                 &  \textasciitilde 4M        & \textasciitilde 39M      & 14           & 140         \\
\cmidrule(lr){2-6}
                          & Chemistry                 & \textasciitilde 4M      & \textasciitilde 30M      & 14           & 140               \\
\cmidrule(lr){2-6}
                          & Material Science                 & \textasciitilde 3M         & \textasciitilde 22M      & 14           & 140            \\
\cmidrule(lr){2-6}
                          & Medicine                 & \textasciitilde 6M & \textasciitilde 30M      & 14           & 140           \\
\cmidrule(lr){2-6}
                          & Physics                 & \textasciitilde 2M        & \textasciitilde 33M         & 14            & 140             \\\midrule
 E-commerce & Amazon      & \textasciitilde 9M       &  \textasciitilde 313M     &  20      &  200              \\\midrule
Literature & Goodreads & \textasciitilde 3M & \textasciitilde 22M & 24 & 240 \\\midrule
Healthcare                & Disease                   & \textasciitilde 47K      & \textasciitilde 4M       & 27           & 270                \\\midrule
Legal          & Freelaw                 & \textasciitilde 84M       & \textasciitilde 114M          & 18            & 180                 \\\midrule
\rowcolor{gray!20}
\textbf{SUM} & - & - & - & \textbf{174} & \textbf{1740} \\\bottomrule
\end{tabular}}
\vspace{-0.25in}
\end{table}

To curate high-quality and diverse data without heavy human effort, the construction of \data contains four steps:
1) We first collect large \textit{reference graph data} from real-world scenarios which can serve as the context for data generation.
2) Then, we manually \textit{design question templates} which can be answered on the reference graph data.
3) After that, we call GPT-4 to generate diverse \textit{question expressions} for each question template.
4) Finally, we \textit{automatically generate ground truth answers} from the domain-specific graphs.

\subsection{Reference Graph Data}


We collect data from five domains where the knowledge lies in the format of graphs: academia, e-commerce, literature, healthcare, and legal.
The detailed statistics of the graphs can be found in Appendix Table \ref{apx-tb-data}.

In the \textbf{academic domain}, papers, authors, and venues are naturally interconnected by citation, ``written-by'', and ``publish-in'' relations. We obtain academic graphs across six disciplines including Biology, Computer Science, Chemistry, Material Science, Medicine, and Physics from DBLP\footnote{\url{https://originalfileserver.aminer.cn/misc/dblp_v14.tar.gz}} \cite{Tang-08KDD} and Microsoft Academic Graph (MAG)\footnote{\url{https://zenodo.org/records/7611544}} \cite{wang2020microsoft,zhang2023effect}. 
Nodes on such graphs are papers, authors, and venues, while edges include citation edges, authorship edges, and venueship edges.

In the \textbf{e-commerce domain}, a single product is assigned a brand, and different products are interlinked through ``also-viewed'' or ``also-bought'' relationships, which naturally embody graph-like structures. 
We use Amazon product datasets\footnote{\url{https://cseweb.ucsd.edu/~jmcauley/datasets/amazon/links.html}} \cite{he2016ups}, which provides the metadata information of items across a myriad of product categories.
Nodes on this graph are items and brands, while edges include ``also-viewed'', ``also-bought'', ``buy-after-viewing'', ``bought-together'', and ``item-brand''.

In the \textbf{literature domain}, the inherent graph structure exists with interconnections between books, authors, publishers, and series. The Goodreads dataset\footnote{\url{https://mengtingwan.github.io/data/goodreads}} \cite{wan2018item} offers an extensive collection of books with their metadata.
Nodes on this graph are books, authors, publishers, and series, while edges include ``written-by'', ``publish-in'', ``book-series'' and so on.

In the \textbf{healthcare domain}, we can construct a graph by considering the diseases with their associated properties. We adopt the biological disease graph Hetionet\footnote{\url{https://github.com/hetio/hetionet}} \cite{himmelstein2017systematic}, which comprehensively summarizes existing disease and their symptoms, with the aim of repurposing drugs.
Nodes on this graph include diseases, symptoms, side effects, compounds, and so on, while edges include ``disease-present-symptom'', ``compound-cause-side effect'' and so on.

In the \textbf{legal domain}, there are rich citation links between cases and opinions (since judges rely on citing opinions from previous cases to write for the current case) which naturally form a graph. We use the data from CourtListener\footnote{\url{https://www.courtlistener.com/}}. 
Nodes on this graph are opinion, opinion-cluster, docket, and court, while edges include ``opinion-citation'', ``opinion-cluster'', ``cluster-docket'', and ``docket-court''.

\subsection{Manually Designed Question Templates}

The question generation phase aims to generate questions that can be answered by LLMs after referring to the domain graphs.
Considering that the generated questions should be accurate and meaningful, we ask four well-trained computer science Ph.D. students to write potential questions that can be answered given the graphs as context.

To comprehensively evaluate the LLMs and their capability to interact with graphs, we ask the annotators to design question templates of three different difficulties:

\begin{itemize}
    \vspace{-0.1in}
    \item \textbf{Easy}: These questions can be answered by looking up the feature/degree of only one node or travel on the graph within one hop. For example, ``What is the price of the \{item\}?'' or ``Who are the authors of \{paper\}?''
    \vspace{-0.1in}
    \item \textbf{Medium}: These questions require reasoning on the graphs for more than one hop and involve returning the feature/degree of nodes. For example, ``Who is the closest collaborator with \{author\} in \{year\}?''.
    \vspace{-0.1in}
    \item \textbf{Hard}: These questions cannot be directly answered by looking up the graph, but the graph can be useful by providing informative context. For example, ``What is the complementary item given this \{query\}?''
    \vspace{-0.1in}
\end{itemize}

It is worth noting that the easy-level and medium-level questions can be answered from the given graph, while the ground truth for hard questions cannot be directly found in the graph. 
All the question templates can be found in Appendix \ref{apx:sec:template}.

Once the question templates are manually designed, we extract values from the graph to transform the templates into actual questions.
For example, given the question template ``How many citations did \{paper\} have in \{year\}?'', we can refer to the academic graphs and sample ``Language Models are Unsupervised Multitask Learners'' as the ``paper'' value and ``2021'' as the ``year'' value. This will result in a real question: ``How many citations did Language Models are Unsupervised Multitask Learners have in 2021?''

\subsection{Diverse Question Expression with GPT-4}

Following the previous steps, we obtain question samples for each graph. However, all samples pertaining to the same template will share the same expressions. For example, inquiring about the price of an item will consistently yield the question ``What is the price of the \{item\}?''. This limits the diversity of the data samples and may lead to a partially comprehensive evaluation.

To this end, we propose to use GPT-4 to paraphrase each question template into five different expressions so that we can have more diverse question samples regarding the same type of question. The prompts for paraphrasing can be found in Appendix \ref{apx:sec:paraphrase-prompt}.

\subsection{Automatic Answer Generation}

The final step is to obtain the ground truth answer from the graph for each generated question.
To achieve this goal, we first implement \textit{graph functions} (\textit{e.g.} neighbor check, degree check), which can be utilized to reason on the graph. Then we implement \textit{function chains} which can serve as a combination of graph functions in order to fetch the ground truth answer from the graph.
The function chains are manually written by annotators for each type of question.
Examples can be found in Appendix \ref{apx:sec:answer}.

\section{Graph Chain-of-Thought}
\begin{figure*}[t]
    \centering
    \includegraphics[width=\textwidth]{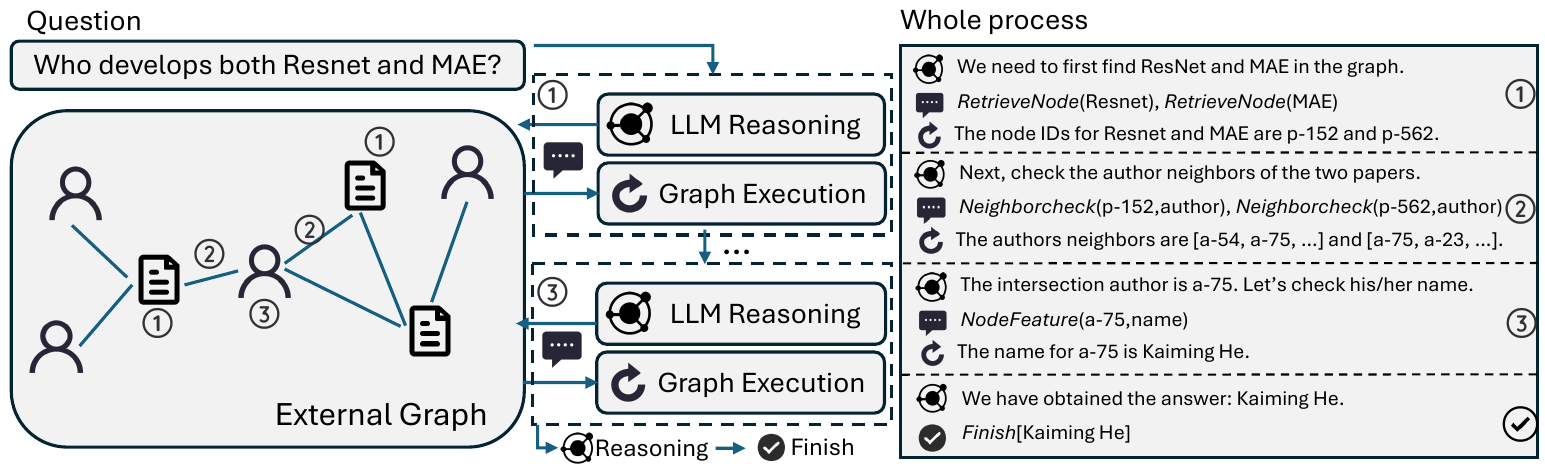}
    \vspace{-0.15in}
    \caption{The workflow of \Ours, an iterative framework with three steps in each iteration: reasoning with LLMs, interaction between LLMs and graphs, and execution on graphs.}\label{fig:graph-cot-overview}
    \vspace{-0.15in}
\end{figure*}


The straightforward solution to let LLMs interact with the graph is through retrieval-augmentation generation (RAG) \cite{lewis2020retrieval, gao2023retrieval}, where a retriever fetches related information from graphs as context for LLM generation.
However, different from text corpus as the external knowledge source, the information in graphs also lies in the complex interconnection between the text units, which poses a potential requirement for traversing and reasoning on graphs. 
To enable LLMs to reason, Chain-of-thought \cite{wei2022chain} is proposed to encourage LLMs to decompose complex tasks into several steps. However, it is designed for reasoning on texts and leaves reasoning on graphs with LLMs an open question.

To this end, we design a simple solution named Graph Chain-of-Thought (\Ours) to tackle the complex graph reasoning problem with LLMs (shown in Figure \ref{fig:graph-cot-overview}).
\Ours is an iterative framework, with three steps in each iteration: \textbf{reasoning}, \textbf{interaction}, and \textbf{execution}.
We delve into each step as follows:

\paragraph{Reasoning with LLMs.}
Given the question or the previous iteration context, the first step is to let the LLMs conduct reasoning on what further external information from graphs is needed to answer the question, or if the question is answerable with the current contexts from graphs.
For example, given the question ``Who are the authors of Language Models are Unsupervised Multitask Learners?''. The LLMs are expected to reason ``We need to first find the paper node \{Language Models are Unsupervised Multitask Learners\} on the graph.''

\paragraph{Interaction between LLMs and Graphs.}
Based on the output results from the previous LLM reasoning step, the next step is to let LLMs know how to interact with the graphs and fetch relevant information from the graphs.
Inspired by~\cite{yao2022react}, we pre-define four
graph functions to cover both the semantic information and structure information on the graphs:
\begin{itemize}
    \vspace{-0.1in}
    \item \textit{RetrieveNode}(Text): Identify related nodes in the graph with semantic search.
    \vspace{-0.1in}
    \item \textit{NodeFeature}(NodeID, FeatureName): Extract the textual feature information from the graph for a specific node.
    \vspace{-0.1in}
    \item \textit{NeighborCheck}(NodeID, NeighborType): Return the neighboring information in the graph for a specific node.
    \vspace{-0.1in}
    \item \textit{NodeDegree}(NodeID, NeighborType): Return the degree of a specific neighbor type for a specific node in the graph.
    \vspace{-0.1in}
\end{itemize}

The task at hand requires LLMs to generate accurate graph function calls, based on their previous reasoning results, to effectively interact with the graph.  In the given example,
the LLMs are expected to generate ``\textit{RetrieveNode}(Language Models are Unsupervised Multitask Learners)''.

\paragraph{Execution on Graphs.}
The final step is to call those functions given by the previous step and fetch the relevant information from the graph.
For the previous example, the graph will execute the \textit{RetrieveNode}($\cdot$) function and return ``The ID of the most relevant paper node is p-4123''.
Then, the process for the current iteration is over, and we start the new iteration from ``reasoning with LLMs''.
The whole framework will be iterated until the LLM finishes the reasoning and outputs the final answer. In this work, we enable LLMs to learn how to conduct \Ours with in-context learning \cite{dong2022survey}. The prompts and demonstrations can be found in Appendix \ref{apx:sec:ours-prompts}.

\paragraph{Connection to LLM agents.}
It is worth mentioning that \Ours can be seen as an agent framework \cite{xi2023rise,zhuang2023toolqa}, where the LLM backbones are the agents and the graphs are the environments. The agents (LLMs) can interact with the environment (graphs) with some predefined functions (defined in this section above). The goal of the agents is to explore the graph environment and conduct question-answering.

\section{Experiments}

\subsection{Experimental Setup}

\begin{table*}[ht]
\centering
\caption{Model performance on \data comparing standard LLMs, text retrieval augmented LLMs (Text RAG), graph retrieval augmented LLMs (Graph RAG), and \Ours. We showcase their performance based on Rouge-L (R-L) and GPT4score. We adopt GPT-3.5-turbo as the backbone for \Ours.}
\setlength{\tabcolsep}{2.5mm}
\scalebox{0.68}{
\begin{tabular}{llcccccccccc}
\toprule
\multicolumn{2}{c}{\multirow{2}{*}{Model}} & \multicolumn{2}{c}{\textbf{Academic}} & \multicolumn{2}{c}{\textbf{E-commerce}} & \multicolumn{2}{c}{\textbf{Literature}} & \multicolumn{2}{c}{\textbf{Healthcare}} & \multicolumn{2}{c}{\textbf{Legal}} \\
\multicolumn{2}{c}{}  & \textbf{R-L} & \textbf{GPT4score} & \textbf{R-L} & \textbf{GPT4score} & \textbf{R-L} & \textbf{GPT4score} & \textbf{R-L} & \textbf{GPT4score} & \textbf{R-L} & \textbf{GPT4score} \\ 
\midrule
\multicolumn{1}{c}{\multirow{3}{*}{\rotatebox[origin=c]{90}{Base}}} & LLaMA-2-13b-chat   & 8.13 &	8.03 & 7.01 	& 12.00  & 5.32 & 20.83 & 5.25 & 13.70   & 15.97 & 16.11     \\
 & Mixtral-8x7b   & 9.02 &	8.14 & 12.54 	& 18.00  & 7.50 & 22.50 & 3.88 & 20.00   & 12.74 & 16.11     \\
& GPT-3.5-turbo   & 6.05 &	12.80 & 9.18 	& 23.50  & 10.43 & 26.67 & 5.83 & 14.44   & 10.51 & 20.00     \\
\midrule
\multicolumn{1}{c}{\multirow{3}{*}{\rotatebox[origin=c]{90}{\parbox[c]{1cm}{\centering Text\\RAG}}}} 
  & LLaMA-2-13b-chat   & 8.69 &	8.52 & 9.23 	& 12.50  & 7.61 & 20.00 & 1.44 & 5.93   & 15.37 & 16.67     \\
& Mixtral-8x7b   & 8.44 &	8.02 & 23.14 	& 29.50  & 13.35 & 27.92 & 3.22 & 16.67   & 19.69 & 25.00     \\
& GPT-3.5-turbo   & 5.83 &	9.91 & 14.06 	& 20.00  & 10.04 & 20.83 & 4.57 & 8.52   & 18.14 & 23.89     \\
\midrule
\multicolumn{1}{c}{\multirow{3}{*}{\rotatebox[origin=c]{90}{\parbox[c]{1cm}{\centering Graph\\RAG}}}} 
  & LLaMA-2-13b   & 22.01 &	22.97 & 12.48 	& 20.00  & 9.25 & 20.00 & 2.97 & 4.81   & 17.98 & 17.22     \\
& Mixtral-8x7b   & 27.77 &	31.20 & 32.87 	& 37.00  & 20.08 & 33.33 & 8.66 & 15.19   & 23.48 & 25.56     \\
& GPT-3.5-turbo   & 18.45 &	26.98 & 17.52 	& 28.00  & 14.94 & 24.17 & 8.69 & 14.07   & 18.66 & 22.22     \\
\midrule
& \Ours  &  \textbf{31.89} & \textbf{33.48} & \textbf{42.40} & \textbf{44.50} & \textbf{41.59} & \textbf{46.25} & \textbf{22.33} & \textbf{28.89} & \textbf{30.52} & \textbf{28.33}            \\
\bottomrule
\end{tabular}
}\label{tab:results}
\end{table*}

\paragraph{Baselines.} We compare our proposed \Ours with three types of baseline methods: standard LLMs (Base LLMs), text retrieval-augmented LLMs (Text RAG LLMs), and graph retrieval-augmented LLMs (Graph RAG LLMs): 
\begin{itemize}
    \vspace{-0.1in}
    \item \textbf{Base LLMs}:  We test if the LLMs can answer the given question with their knowledge without interacting with external data.
    We adopt the standard prompting, which involves providing simple instructions and letting LLMs generate an answer for the question.
    \vspace{-0.1in}
    \item \textbf{Text RAG LLMs} \cite{gao2023retrieval}: We treat the external graphs as pure text corpora and utilize a retriever to retrieve relevant text information from them.
    Subsequently, the retrieved text serves as context to augment the LLM for question answering.
    \vspace{-0.1in}
    \item \textbf{Graph RAG LLMs}: This is an extension of text RAG, where not only the retrieved text/node but also the subgraph associated with it is linearized into a text sequence \cite{ye2023natural} and serves as the context. In the main result, we use 1-hop ego-graphs.
    \vspace{-0.1in}
\end{itemize}
For all categories of baselines, we explore three LLM backbones, including LLaMA-2-13b-chat \cite{touvron2023llama}, Mixtral-8x7b-Instruct \cite{albert2024mixtral}, and GPT-3.5-turbo \cite{ouyang2022training}.
R-L

\paragraph{Evaluation Metrics.}
We use both rule-based metrics and model-based metrics to comprehensively evaluate the model results. 
For the former, we use Rouge-L(R-L), which measures the longest common subsequence of words between the responses and the ground truth answers. 
For the latter, we call GPT-4 to measure if the model output and ground truth are the same. We calculate the percentage of ``correct'' predicted by GPT-4 as GPT4score.

\paragraph{Implementation Settings.}
All experiments are conducted on NVIDIA GeForce RTX A6000 GPUs with Python 3.8 and Huggingface 4.36.2.
We use Mpnet-v2 \footnote{\url{https://huggingface.co/sentence-transformers/all-mpnet-base-v2}} as the retriever for all the baselines and our method and implement the indexing with FAISS \cite{johnson2019billion}.
In \Ours, we adopt GPT-3.5-turbo-16k (Jan 2024) as the backbone LLM in the main results and set the temperature $t$ to 0 for consistent responses. We provide demonstrations for \Ours on how to conduct reasoning in Appendix \ref{apx:sec:ours-prompts}.

\vspace{-0.1in}
\subsection{Overall Performance}
\vspace{-0.05in}
The main results are shown in Table \ref{tab:results}. From the results, we can find that:
1) \Ours outperforms all the baselines consistently and significantly.
2) Base LLMs are exhibiting fairly poor performance, typically because the LLMs may not contain the knowledge needed to answer those questions.
3) Graph RAG LLMs outperform text RAG LLMs in most cases since the former can provide more structure-aware context, which is helpful for problem-solving.
4) While \Ours performs the best, the absolute score is not high, leaving a great space to improve.

\subsection{Ablation Study}
\vspace{-0.05in}
\begin{figure}[t]
\centering
\includegraphics[width=0.46\textwidth]{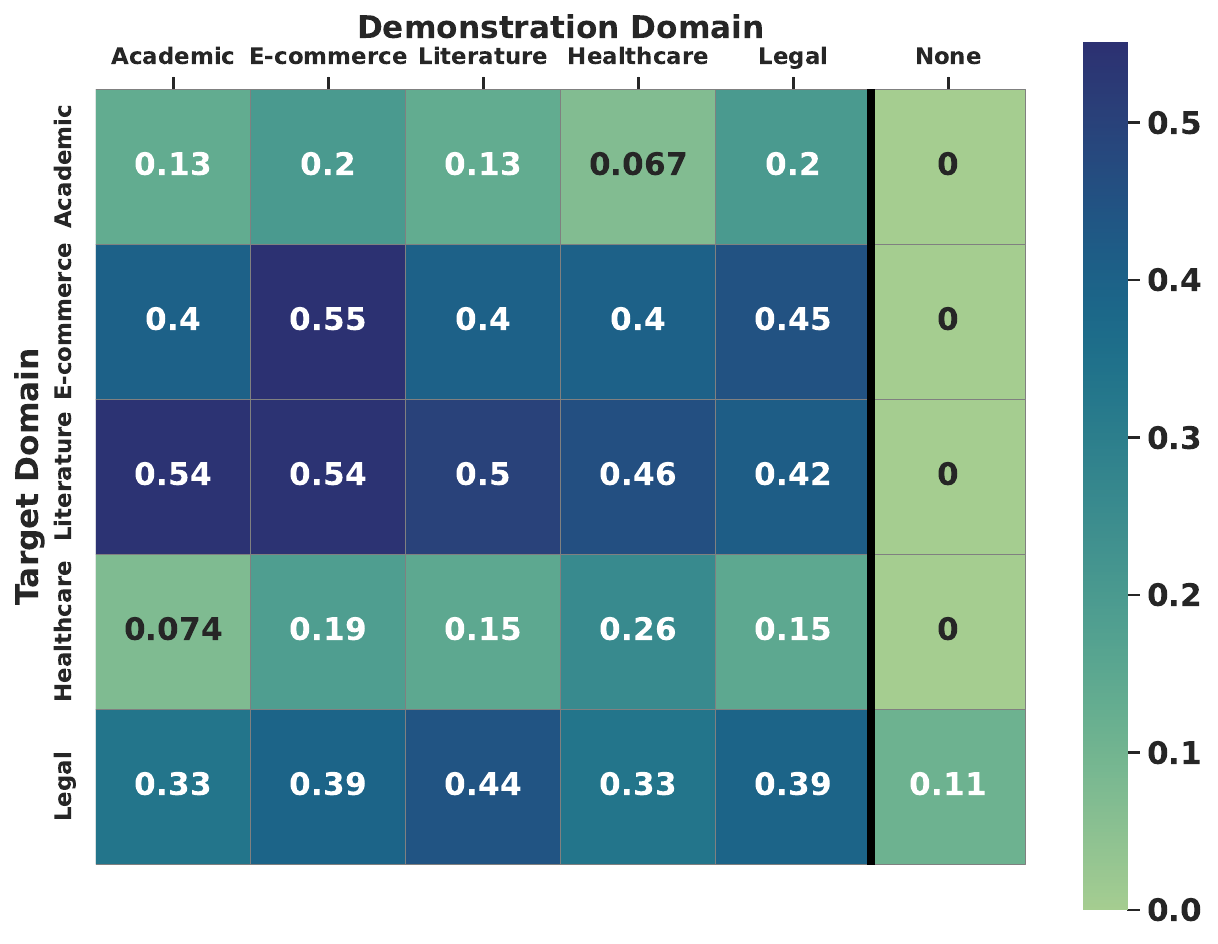}
\vspace{-0.15in}
\caption{Ablation study of \Ours.
It performs well with in-domain demonstrations and remains generally robust to domain shifts in demonstrations.
}\label{fig:cross-domain}
\vspace{-0.15in}
\end{figure}

\paragraph{How Important are the Demonstrations for \Ours?}
To answer this question, we conduct experiments from two aspects: zero-shot study (no demonstrations) and cross-domain study (demonstrations from other domains \cite{ding2018graph}).
The results are shown in Figure \ref{fig:cross-domain}, where the columns and rows correspond to the source domain and target domain respectively.
For the zero-shot study, no demonstrations are given (rightest column in Figure \ref{fig:cross-domain}). We empirically find that given no reasoning demonstrations, \Ours cannot work in all the datasets (nearly 0 performance). This implies that the LLMs suffer if given insufficient instructions (only graph definition and interaction function definitions).
For the cross-domain study, we provide demonstrations from the source domain graphs and test on the target domain graphs. From the result (left five columns in Figure \ref{fig:cross-domain}), in-domain demonstrations (diagonal) perform quite well and \Ours is overall robust to demonstration domain-shift. This observation underscores the adaptability and effectiveness of \Ours in capturing the key steps of graph chain-reasoning through in-context learning, despite the diverse demonstration domains.


\paragraph{How Different LLMs Perform in \Ours?}
In the main results, we adopt GPT-3.5-turbo as the LLM backbone for \Ours. In this section, we explore \Ours with other LLM backbones including LLaMA-2-13b-chat, Mixtral-8x7b-Instruct, GPT-3.5-turbo, and GPT-4. We randomly extract a subset from \data (one sample for each question template) to experiment and the results are shown in Table \ref{tab:backbone-results}.
From the result, we find that the LLM backbone matters. An LLM with more advanced instruction following ability and reasoning ability (\textit{i.e.}, GPT-4) can contribute to better performance in \Ours.

\begin{table}[t]
\centering
\caption{Results of \Ours with different LLM backbones.}
\setlength{\tabcolsep}{2.5mm}
\scalebox{0.9}{
\begin{tabular}{lc}
\toprule
Model  & \textbf{GPT4score}   \\ 
\midrule
\Ours     \\
 \quad w. LLaMA-2-13b-chat  &	16.04    \\
 \quad w. Mixtral-8x7b    & 36.46     \\
 \quad w. GPT-3.5-turbo    &	36.63     \\
 \quad w. GPT-4   &	46.28   \\
\bottomrule
\end{tabular}
}
\vspace{-0.1in}
\label{tab:backbone-results}
\end{table}

\subsection{RAG vs \Ours}
\paragraph{Is the Retrieval-Augmented LLM a Good Choice on Graphs?}
We study how graph retrieval-augmented LLMs work by setting the retrieved subgraph to be just one node, 1-hop ego-graphs, and 2-hop ego-graphs. For all the settings, the ego-graphs are linearized into text sequences and serve as context. The averaged results over all the datasets are shown in Table \ref{tab:rag-results}. From the results, retrieving 1-hop ego-graph performs the best, but still underperforms \Ours. The reason is that when doing subgraph retrieval, the number of nodes/texts will grow exponentially as the hop number grows linearly. Even though the bigger the subgraph is, the more information it contains, a large-hop ego-graph will lead to a super long context which is even over the maximum input length of LLMs and will cause LLMs to lose in the middle. In this case, \Ours can serve as a better way to extract more useful information from the graph.


\begin{table}[t]
\centering
\caption{Results of LLM with different retrieval-augmentation methods on \data.}
\setlength{\tabcolsep}{2.5mm}
\scalebox{0.9}{
\begin{tabular}{lc}
\toprule
Model & \textbf{GPT4score}   \\ 
\midrule
 GPT-3.5-turbo   &	19.48     \\
 + node retrieval    &	16.63    \\
 + 1-hop subgraph retrieval   &	\textit{23.09}     \\
 + 2-hop subgraph retrieval  &	22.12     \\
 + \Ours    &	\textbf{36.29}     \\
\bottomrule
\end{tabular}
}\label{tab:rag-results}
\end{table}

\subsection{\Ours on Questions of Different Difficulty Levels in \data}
\begin{figure}[t]
\centering
\includegraphics[width=0.48\textwidth]{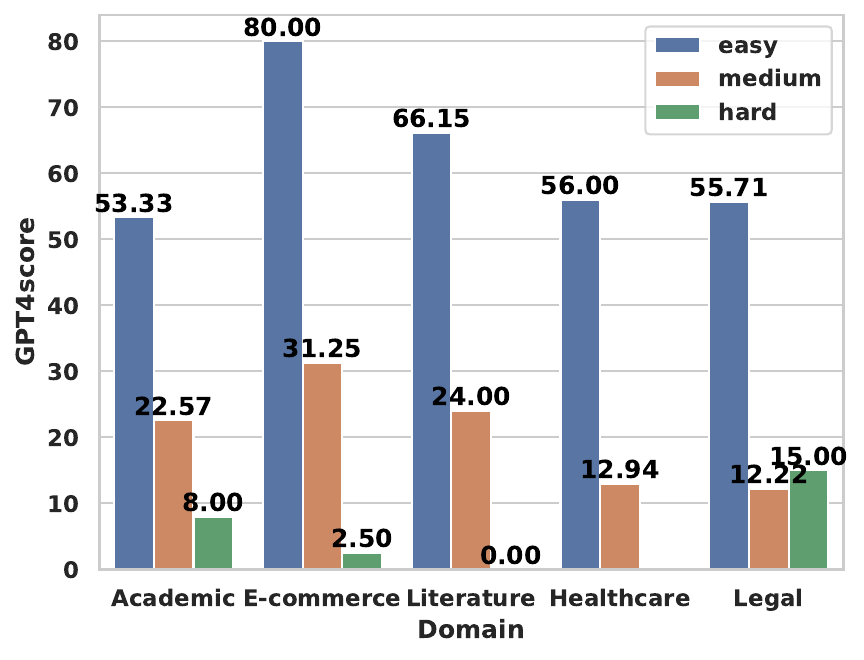}
\caption{Results of \Ours on samples of different difficulties in \data.}\label{fig:q-level}
\end{figure}

In this section, we analyze the performance of \Ours on questions of different difficulty levels. The results are shown in Figure \ref{fig:q-level}, where we find that \Ours performs relatively high on easy questions (the reasoning chains for those questions are simple) while having worse performance on medium/hard questions (complex/inductive reasoning).

\subsection{Case Studies of \Ours}
\begin{figure*}[t]
    \centering
    \includegraphics[width=\textwidth]{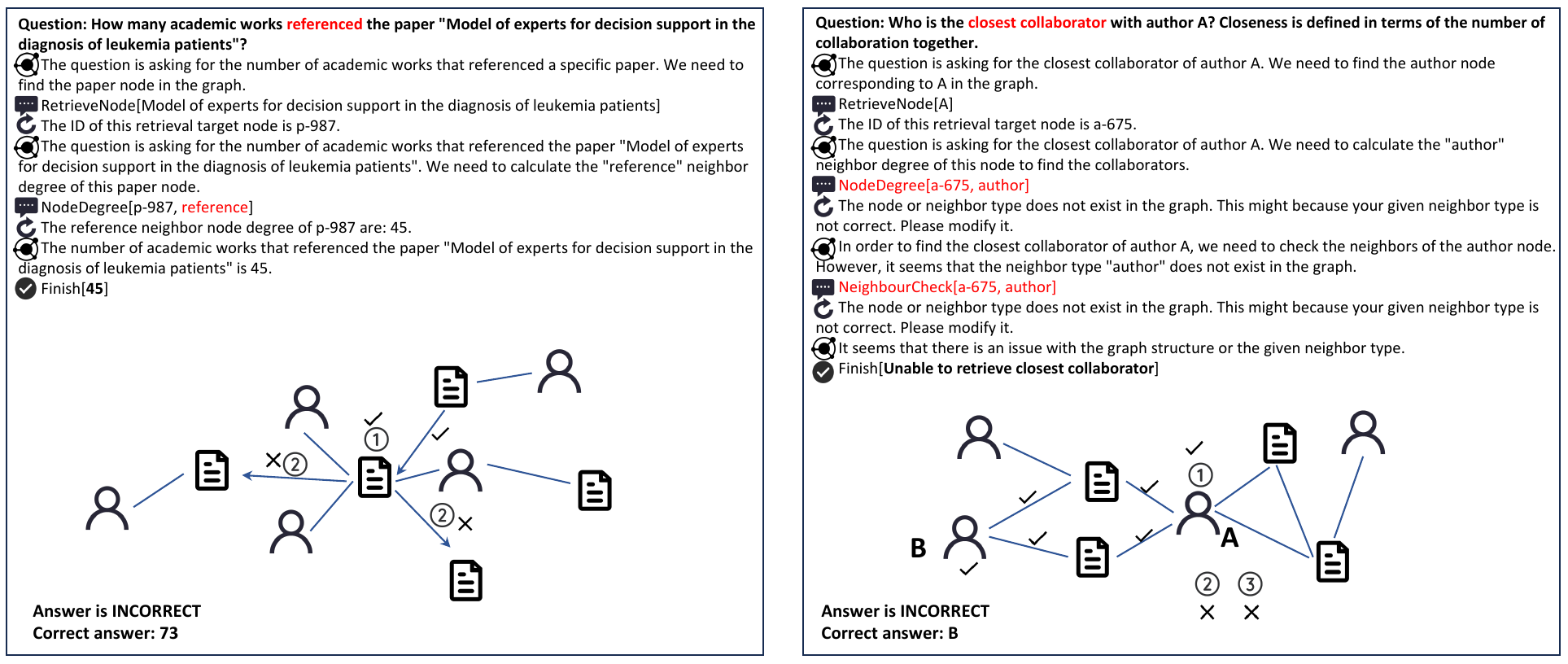}
    \caption{Failure cases of \Ours. The key information in the question and the wrong interaction of \Ours are colored in \textcolor{red}{red}. The author names in the second example are anonymized as A \& B.}\label{fig:graph-cot-example}
\end{figure*}

We conduct case studies to understand the weakness of \Ours. The results of two failure cases are shown in Figure \ref{fig:graph-cot-example}. For the left case, we can find that despite using the most advanced LLM backbone (\textit{i.e.}, GPT-4), the framework sometimes refers to the occurrence of the word rather than understanding its semantic meaning, leading to the wrong interaction function calls. For the right case, we can find that the framework sometimes misunderstands the structure of the graph, resulting in interaction failures.

Although \Ours achieves relatively good performance on \data, there is still quite some room to improve. The two most promising directions to enhance LLMs' graph reasoning abilities are exploring \textit{how to let LLMs better understand the graphs} and \textit{how to let LLMs conduct more complex reasoning}.
For the former, in \Ours, we mainly use natural language to describe the graph for LLMs. Given that graphs are more structured rather than sequential, more structure-aware languages (\textit{e.g.}, graphXML \cite{herman2000graphxml}) can be better choices.
For the latter, given that reasoning problems on graphs are not only chain-reasoning problems, some more advanced reasoning paradigms such as tree-based reasoning \cite{yao2023tree} and graph-based reasoning \cite{besta2023graph} can be good directions.

\section{Related Work}
\subsection{LLMs on graphs}
Inspired by the recent success of LLMs on natural language processing tasks, researchers are exploring solving graph tasks with LLMs \cite{jin2023large}. 
The main idea is to serve LLMs as the feature extractor \cite{chen2023exploring} or final predictor \cite{jin2023patton}.
For the former, many methods adopt a LLM-GNN cascaded structure \cite{chien2021node}, where LLMs extract node features for graph neural networks (GNNs) \cite{wu2020comprehensive}. For example, SimTeG \cite{duan2023simteg} proposes to first warm up the LLM feature extractor before training the whole pipeline. GLEM \cite{zhao2022learning} introduces an iterative pipeline where GNNs can provide feedback for LLM feature extractors.
For the latter, existing works transfer the structure information into a sequence to feed into LLMs \cite{tian2023graph, xiong2024large} or design advanced graph-empowered LLMs \cite{yang2021graphformers}. For example, InstructGLM \cite{ye2023natural} utilizes natural language to describe graph structure. Heterformer \cite{jin2023heterformer} proposes a graph-nested language model architecture.
However, most existing works mainly focus on traditional graph tasks such as node classification \cite{xiao2022graph} and link prediction \cite{zhang2018link}. 
On the other hand, Graph-of-thought \cite{besta2023graph} proposes to conduct LLM reasoning with graph-structured thinking. Nevertheless, it mainly focuses on text-based reasoning rather than referring to external graphs.
In our work, we research the question of augmenting LLMs with external graphs by conducting graph reasoning with LLMs.


\subsection{Augmenting LLMs with external knowledge}
Although LLMs \cite{touvron2023llama,albert2024mixtral} have shown their superb language understanding and generation capability \cite{zhao2023survey}, they encounter issues with generating misleading information that seems credible but lacks factual basis, a phenomenon known as hallucination \cite{tonmoy2024comprehensive, rawte2023survey}. To alleviate such an issue, existing works \cite{shuster2021retrieval} propose to augment LLMs with text corpora as external knowledge sources, with the retrieval-augmentation framework proposed \cite{lewis2020retrieval, gao2023retrieval}. Before LLMs' inference, relevant text units are retrieved from the corpora \cite{karpukhin2020dense} and serve as the context for LLMs to help reduce hallucination \cite{dong2022survey}. \citet{lewis2020retrieval} proposes to train the whole framework with a retriever and a generator end-to-end. \citet{izacard2020leveraging} introduces a fusion-in-decoder architecture to jointly consider all retrieved contexts in the generation.
However, most existing works are designed to utilize external text corpora to augment LLMs.
In our work, we explore how to augment LLMs with external text-attributed graphs and propose a benchmark for evaluation.


\section{Conclusions}
In this work, we study the problem of augmenting LLMs with (text-attributed) graphs as external knowledge sources.
We first manually construct a benchmark dataset called \data, which contains 1,740 questions and 10 graphs from 5 domains. Each question in \data can be answered by referring to the graphs.
We further propose a simple and effective framework called \Ours, which can augment LLMs with graphs by letting LLMs conduct iterative reasoning on graphs. \Ours contains three sub-steps in each iteration: LLM reasoning, LLM-graph interaction, and graph execution.
We then conduct experiments with three backbone LLMs on \data and demonstrate the effectiveness of \Ours.
Future works can explore how to let LLMs better understand the graphs and how to let LLMs conduct more complex reasoning.

\section*{Limitations}

In this work, we mainly focus on augmenting LLMs with external graphs as knowledge sources by reasoning on the graphs, with a comprehensive benchmark dataset proposed.
For \data construction, although we used GPT-4 to paraphrase the question templates, they are still mostly designed manually, so there might be room for improvement in terms of question diversity and difficulty. 
For \Ours, the LLM backbone used is an API model that cannot be fine-tuned (or is very costly to fine-tune). Future methods might need to consider how to train the LLMs explicitly to navigate on graphs.

\section*{Ethics Statement}
Research has demonstrated the proficiency of Large Language Models (LLMs) \cite{touvron2023llama,albert2024mixtral} in mastering language processing and generation. However, investigations have also pointed out their limitations, including social biases \cite{liang2021towards} and the propagation of false information \cite{abid2021persistent}. Our study aims to enhance LLMs by integrating external graphs as knowledge sources, proposing this approach as a potential solution to reduce bias and eradicate misinformation.

\section*{Acknowledgements}
We thank Chen Yan (J.D.) for providing legal domain knowledge to help the authors construct the legal graph.
Research was supported in part by US DARPA KAIROS Program No. FA8750-19-2-1004 and INCAS Program No. HR001121C0165, National Science Foundation IIS-19-56151, and the Molecule Maker Lab Institute: An AI Research Institutes program supported by NSF under Award No. 2019897, and the Institute for Geospatial Understanding through an Integrative Discovery Environment (I-GUIDE) by NSF under Award No. 2118329. Any opinions, findings, and conclusions or recommendations expressed herein are those of the authors and do not necessarily represent the views, either expressed or implied, of DARPA or the U.S. Government. The views and conclusions contained in this paper are those of the authors and should not be interpreted as representing any funding agencies.

\bibliography{custom}

\newpage

\appendix

\section{Dataset}\label{apx:sec:data}

The detailed statistics of the graphs in \data are shown in Table \ref{apx-tb-data}. We will discuss the nodes' features and neighbors in each graph in the following paragraphs respectively.

\noindent\textbf{Academic Graphs} contain three types of nodes: paper, author, and venue. Here are examples to show the feature information and neighboring information for the three types of nodes respectively.



\begin{lstlisting}[language=json]
# paper node
{
    'features': {
        'title': ..., 
        'abstract': ..., 
        'keywords': [...], 
        'lang': ..., 
        'year': ...,
        }, 
    'neighbors': {
        'author': [...], 
        'venue': [...], 
        'reference': [...], 
        'cited_by': [...],
        }
}
# author node
{
    'features': {
        'name': ..., 
        'organization': ..., 
        }, 
    'neighbors': {
        'paper': [...], 
        }
}
# venue node
{
    'features': {
        'name': ..., 
        }, 
    'neighbors': {
        'paper': [...], 
        }
}
\end{lstlisting}

\vspace{0.1in}
\noindent\textbf{E-commerce Graph} contains two types of nodes: item and brand. Here are examples to show the feature information and neighboring information for the two types of nodes respectively.

\begin{lstlisting}[language=json]
# item node
{
    'features': {
        'title': ..., 
        'description': ..., 
        'price': ..., 
        'category': [...], 
        }, 
    'neighbors': {
        'also_viewed_item': [...], 
        'buy_after_viewing_item': [...], 
        'also_bought_item': [...], 
        'bought_together_item': [...],
        'brand': [...],
        }
}
# brand node
{
    'features': {
        'name': ..., 
        }, 
    'neighbors': {
        'item': [...], 
        }
}
\end{lstlisting}

\vspace{0.1in}
\noindent\textbf{Literature Graph} contains four types of nodes: book, author, publisher, and series. Here are examples to show the feature information and neighboring information for the four types of nodes respectively.

\begin{lstlisting}[language=json]
# book node
{
    'features': {
        'country_code': ..., 
        'language_code': ..., 
        'is_ebook': ..., 
        'title': ..., 
        'description': ..., 
        'format': ..., 
        'num_pages': ..., 
        'publication_year': ..., 
        'popular_shelves': [...], 
        'genres': [...], 
        }, 
    'neighbors': {
        'author': [...], 
        'publisher': [...], 
        'series': [...], 
        'similar_books': [...],
        }
}
# author node
{
    'features': {
        'name': ..., 
        }, 
    'neighbors': {
        'book': [...], 
        }
}
# publisher node
{
    'features': {
        'name': ..., 
        }, 
    'neighbors': {
        'book': [...], 
        }
}
# series node
{
    'features': {
        'title': ..., 
        'description': ..., 
        }, 
    'neighbors': {
        'book': [...], 
        }
}
\end{lstlisting}

\vspace{0.1in}
\noindent\textbf{Healthcare Graph} contains eleven types of nodes: anatomy, biological process, cellular component, compound, disease, gene, molecular function, pathway, pharmacologic class, side effect, and symptom. Here are examples to show the feature information and neighboring information for the eleven types of nodes respectively.

\begin{lstlisting}[language=json]
# anatomy node
{
    'features': {
        'name': ..., 
        }, 
    'neighbors': {
        'Anatomy-expresses-Gene': [...], 
        }
}
# biological process node
{
    'features': {
        'name': ..., 
        }, 
    'neighbors': {
        'Gene-participates-Biological Process': [...], 
        }
}
# cellular component node
{
    'features': {
        'name': ..., 
        }, 
    'neighbors': {
        'Gene-participates-Cellular Component': [...], 
        }
}
# compound node
{
    'features': {
        'name': ..., 
        }, 
    'neighbors': {
        'Compound-causes-Side Effect': [...],
        'Compound-upregulates-Gene': [...],
        'Compound-downregulates-Gene': [...],
        }
}
# disease node
{
    'features': {
        'name': ..., 
        }, 
    'neighbors': {
        'Disease-associates-Gene': [...],
        'Disease-localizes-Anatomy': [...],
        'Compound-treats-Disease': [...],
        'Disease-resembles-Disease': [...],
        'Disease-presents-Symptom': [...],
        'Disease-upregulates-Gene': [...],
        }
}
# gene node
{
    'features': {
        'name': ..., 
        }, 
    'neighbors': {
        'Gene-participates-Biological Process': [...],
        'Anatomy-upregulates-Gene': [...],
        'Anatomy-expresses-Gene': [...],
        'Anatomy-downregulates-Gene': [...],
        'Compound-upregulates-Gene': [...],
        'Gene-interacts-Gene': [...],
        'Gene-participates-Molecular Function': [...],
        'Gene-participates-Cellular Component': [...],
        }
}
# molecular function node
{
    'features': {
        'name': ..., 
        }, 
    'neighbors': {
        'Gene-participates-Molecular Function': [...],
        }
}
# pathway node
{
    'features': {
        'name': ..., 
        }, 
    'neighbors': {
        'Gene-participates-Pathway': [...],
        }
}
# pharmacologic node
{
    'features': {
        'name': ..., 
        }, 
    'neighbors': {
        'Pharmacologic Class-includes-Compound': [...],
        }
}
# side effect node
{
    'features': {
        'name': ..., 
        }, 
    'neighbors': {
        'Compound-causes-Side Effect': [...],
        }
}
# symptom node
{
    'features': {
        'name': ..., 
        }, 
    'neighbors': {
        'Disease-presents-Symptom': [...],
        }
}
\end{lstlisting}

\vspace{0.1in}
\noindent\textbf{Legal Graph} contains four types of nodes: opinion, opinion cluster, docket, and court. Here are examples to show the feature information and neighboring information for the four types of nodes respectively.

\begin{lstlisting}[language=json]
# opinion node
{
    'features': {
        'plain_text': ..., 
        }, 
    'neighbors': {
        'opinion_cluster': [...], 
        'reference': [...], 
        'cited_by': [...], 
        }
}
# opinion cluster node
{
    'features': {
        'judges': ..., 
        'case_name': ..., 
        'attorneys': ..., 
        'syllabus': ..., 
        }, 
    'neighbors': {
        'opinion': [...], 
        'docket': [...], 
        }
}
# docket node
{
    'features': {
        'case_name': ..., 
        'pacer_case_id': ..., 
        }, 
    'neighbors': {
        'opinion_cluster': [...], 
        'court': [...], 
        }
}
# court node
{
    'features': {
        'citation_string': ..., 
        'full_name': ..., 
        'start_date': ..., 
        'end_date': ..., 
        }, 
    'neighbors': {
        'docket': [...], 
        }
}
\end{lstlisting}

\begin{table*}[t]
\centering 
\caption{Detailed Dataset Statistics of \data.}\label{apx-tb-data}
\renewcommand\arraystretch{0.88}
\fontsize{8}{10}\selectfont \setlength{\tabcolsep}{0.4em}
\scalebox{1.1}{
\begin{tabular}{@{}cccc cc@{}}
\toprule
\multirow{2}{*}{Domain}   & \multirow{2}{*}{Topic} & \multicolumn{2}{c}{Graph Statistics} & \multicolumn{2}{c}{Data}      \\ 
\cmidrule(lr){3-4} \cmidrule(lr){5-6} 
                          &                        & Nodes      & Edges         & \# Templates & \# Questions  \\ \midrule
\multirow{18}{*}{Academic} & \multirow{3}{*}{CS}                 & Paper (\textasciitilde 5M)         & Written-by (\textasciitilde 14M) &   \multirow{3}{*}{15}      &   \multirow{3}{*}{150}      \\
&  &  Author (\textasciitilde 2M) & Publish-in (\textasciitilde 5M)   \\
&  &  Venue (\textasciitilde 55K) & Cited-by (\textasciitilde 32M)      \\
\cmidrule(lr){2-6}
                          &  \multirow{3}{*}{Biology}                & Paper (\textasciitilde 1M)          & written-by (\textasciitilde 8M) &   \multirow{3}{*}{14}      &   \multirow{3}{*}{140}    \\
&  &  Author (\textasciitilde 2M) & publish-in (\textasciitilde 1M)     \\
&  &  Venue (100) & cited-by (\textasciitilde 29M)     \\
\cmidrule(lr){2-6}
                          &  \multirow{3}{*}{Chemistry}                & Paper (\textasciitilde 1M)          & written-by (\textasciitilde 7M)     &   \multirow{3}{*}{14}      &   \multirow{3}{*}{140}    \\
&  &  Author (\textasciitilde 2M) & publish-in (\textasciitilde 1M)    \\
&  &  Venue (100) & cited-by (\textasciitilde 20M)        \\
\cmidrule(lr){2-6}
                          &  \multirow{3}{*}{Material Science}                & Paper (\textasciitilde 1M)          & written-by (\textasciitilde 6M)    &   \multirow{3}{*}{14}      &   \multirow{3}{*}{140}       \\
&  &  Author (\textasciitilde 1M) & publish-in (\textasciitilde 1M)         \\
&  &  Venue (99) & cited-by (\textasciitilde 14M)        \\
\cmidrule(lr){2-6}
                          &  \multirow{3}{*}{Medicine}                & Paper (\textasciitilde 2M)          & written-by (\textasciitilde 14M)      &   \multirow{3}{*}{14}      &   \multirow{3}{*}{140}       \\
&  &  Author (\textasciitilde 4M) & publish-in (\textasciitilde 2M)          \\
&  &  Venue (100) & cited-by (\textasciitilde 12M)              \\
\cmidrule(lr){2-6}
                          &  \multirow{3}{*}{Physics}                & Paper (\textasciitilde 1M)          & written-by (\textasciitilde 13M) &   \multirow{3}{*}{14}      &   \multirow{3}{*}{140}        \\
&  &  Author (\textasciitilde 1M) & publish-in (\textasciitilde 1M)          \\
&  &  Venue (91) & cited-by (\textasciitilde 18M)           \\
\midrule
 \multirow{5}{*}{E-commerce} & \multirow{5}{*}{Amazon}      & \multirow{2}{*}{Item (\textasciitilde 9M)}       &  also-viewed (\textasciitilde 125M)     &   \multirow{5}{*}{20}      &   \multirow{5}{*}{200}              \\
 &    &  &  buy-after-viewing (\textasciitilde 9M)     \\
 &    &  \multirow{3}{*}{Brand (\textasciitilde 110K)} & also-bought (\textasciitilde 170M)   \\
 &    &  &  bought-together (\textasciitilde 6M)  \\
 &    &  &  item-brand (\textasciitilde 1M)   \\
 \midrule
\multirow{4}{*}{Literature} & \multirow{4}{*}{Goodreads} & Book (\textasciitilde 2M) & written-by (\textasciitilde 3M) & \multirow{4}{*}{24} & \multirow{4}{*}{240} \\
 &  &  Author (\textasciitilde 829K) & published-in (\textasciitilde 1M) \\
 &  &  Publisher (\textasciitilde 193K)  &  book-series (\textasciitilde 822K) \\
 &  &  Series (\textasciitilde 400K)  & similar-book (\textasciitilde 16M) \\
\midrule
\multirow{2}{*}{Healthcare}               & \multirow{2}{*}{Disease}          & 11 nodes types      & 24 edge types       & \multirow{2}{*}{27}           & \multirow{2}{*}{270}                \\
 &  &  See Sec \ref{apx:sec:data}  &  See Sec \ref{apx:sec:data} & &  \\
\midrule
\multirow{4}{*}{Legal} & \multirow{4}{*}{Freelaw} & Opinion (\textasciitilde 9M) & opinion-cluster (\textasciitilde 9M) & \multirow{4}{*}{18} & \multirow{4}{*}{180} \\
 &  &  Opinion-cluster (\textasciitilde 8M) & opinion-citation (\textasciitilde 29M) \\
 &  &  Docket (\textasciitilde 66M)  &  cluster-docket (\textasciitilde 8M) \\
 &  &  Court (\textasciitilde 3K)  & docket-court (\textasciitilde 66M) \\
\midrule
\rowcolor{gray!20}
\textbf{SUM} & - & - & - & \textbf{174} & \textbf{1740} \\\bottomrule
\end{tabular}}
\end{table*}

\section{Question Templates}\label{apx:sec:template}
In this section, we show the templates for easy, medium, and hard questions.

\paragraph{Academic Graphs.}
\begin{itemize}
    \item \textbf{Easy}
    \begin{itemize}
        \item \textit{Who are the authors of paper "\{paper\_title\}?"} 
        \item \textit{What organization is researcher \{author\_name\} affiliated with?}
        \item \textit{Where is the paper "\{paper\_title\}" published?}
        \item \textit{How many papers cite the paper "\{paper\_title\}"?}
        \item \textit{How many papers does paper "\{paper\_title\}" cite?}
        \item \textit{Which is the most cited paper by author \{author\_name\} in \{org\_name\}?}
        \item \textit{How many papers did author \{author\_name\} in \{org\_name\} write?}
    \end{itemize}
    \item \textbf{Medium}
        \begin{itemize}
            \item \textit{Who collaborate with author \{author\_name\} in \{org\_name\} to write paper "\{paper\_title\}"?}
            \item \textit{Who wrote both the paper "\{paper1\_title\}" and paper "\{paper2\_title\}"?}
            \item \textit{Who is the closest collaborator with author \{author\_name\} in \{org\_name\}? Closeness is defined in terms of the number of collaborations together.}
            \item \textit{How many collaborators does author \{author\_name\} in \{org\_name\} have in {year}?}
            \item \textit{How many papers did \{author\_name1\} in \{org\_name1\} and \{author\_name2\} in \{org\_name2\} write together?}
            \item \textit{Which venue did \{author\_name1\} in \{org\_name1\} and \{author\_name2\} in \{org\_name2\} collaborate most?}
            \item \textit{How many people does author \{author\_name1\} in \{org\_name1\} need to know at least to know author \{author\_name2\} in \{org\_name2\}?"}
            \item \textit{What is the research interests (top 3 keywords) of author \{author\_name\} in \{org\_name\}?}
        \end{itemize}
    \item \textbf{Hard}
        \begin{itemize}
            \item \textit{Which paper should be recommended to the reader of paper \{paper1\_title\}? Please select from the candidate list \{paper2\_title\}, \{paper3\_title\}, \{paper4\_title\}, \{paper5\_title\}, \{paper6\_title\}, \{paper7\_title\}, \{paper8\_title\}, \{paper9\_title\}, \{paper10\_title\}, \{paper11\_title\}. Please answer the paper title rather than ID."}
        \end{itemize}
\end{itemize}

\paragraph{E-commerce Graph.}
\begin{itemize}
    \item \textbf{Easy}
    \begin{itemize}
        \item \textit{What is the brand of item \{item\_title\}?}
        \item \textit{What is the category of item \{item\_title\}?}
        \item \textit{What is the price of item \{item\_title\}?}
    \end{itemize}
    \item \textbf{Medium}
        \begin{itemize}
            \item \textit{How many co-viewed items does item \{item\_title\} have?}
            \item \textit{How many bought-together items does item \{item\_title\} have?}
            \item \textit{How many buy-after-viewing items does item \{item\_title\} have?}
             \item \textit{How many also-bought items does item \{item\_title\} have?}
             \item \textit{How many items are in brand \{brand\_name\}?}
             \item \textit{Find the items which are in the same brand and same category as item \{item\_title\}.}
            \item \textit{Which item shares over \{num\} co-viewed items with item \{item\_title\}?}
            \item \textit{Which item shares over \{num\} bought-together items with item \{item\_title\}?}
            \item \textit{How many items have the same bought-together items with item \{item\_title\}?}
            \item \textit{What is the average price of the bought-together/co-viewed items with \{item\_title\}?}
            \item \textit{What is the most popular category name of the bought-together/co-viewed items with \{item\_title\}?}
        \end{itemize}
    \item \textbf{Hard}
        \begin{itemize}
            \item \textit{What next item should be recommended to the user based on his history: \{item\_titles\}?}
            \item \textit{What is the exact matched item given this query: \{query\_text\}? }
            \item \textit{What is the substitutive item given this query: \{query\_text\}? }
           \item \textit{What is the complementary item given this query: \{query\_text\}? }
        \end{itemize}
\end{itemize}

\paragraph{Literature Graph.}
\begin{itemize}
    \item \textbf{Easy}
    \begin{itemize}
        \item \textit{Who are the authors of book \{book title\}?}
        \item \textit{What is the publisher of book \{book title\}?}
        \item \textit{Which shelves do we need to put book \{book title\} on?}
        \item \textit{What genre does the book \{book title\} belong to?}
        \item \textit{In which series is the book \{book title\} included?}
        \item \textit{What is the publication year of book \{book title\}?}
        \item \textit{How many pages does the book \{book title\} have?}
        \item \textit{Is the book \{book title\} an eBook?}
        \item \textit{What language is the book \{book title\} written in?}
        \item \textit{How many books has author \{author name\} written?}
        \item \textit{How many similar books does Book \{book title\} have?}
        \item \textit{How many books does publisher \{publisher name\} publish?}
        \item \textit{How many books are part of the series \{series title\}?}
    \end{itemize}
    \item \textbf{Medium}
        \begin{itemize}
            \item \textit{Find the book written by the same author and published by the same publisher as book \{book title\}.}
            \item \textit{Find books by the same author and share similar genre with book \{book title\}.}
            \item \textit{Find the earliest book written by the author of the book \{book title\}.}
            \item \textit{Find the series in which the same author as the book \{book title\} has contributed, but the series is different from the book's series.}
            \item \textit{How many authors have collaborated with the publisher \{publisher name\}?}
            \item \textit{Which author has the most published books that have the same genre as the book \{book title\}?}
            \item \textit{What is the most common publication format of books by author \{author name\}?}
            \item \textit{What is the most frequent genre in the works of the author \{author name\}?}
            \item \textit{Which publisher has released the majority of books in the genre \{genre name\}?}
            \item \textit{What is the most common language among the books written by author \{author name\}?}
        \end{itemize}
    \item \textbf{Hard}
        \begin{itemize}
            \item \textit{What book should be recommended to the user based on his history: \{book titles\}?}
        \end{itemize}
\end{itemize}

\paragraph{Healthcare Graph.}
\begin{itemize}
    \item \textbf{Easy}
    \begin{itemize}
        \item \textit{What are the side effects of compound \{compound name\}?}
        \item \textit{What are the symptoms of the disease \{disease name\}?}
        \item \textit{What are the biological processes of gene \{gene name\}?}
        \item \textit{What are the molecular functions of gene \{gene name\}?}
        \item \textit{What anatomy can be downregulated by gene \{gene name\}?}
        \item \textit{What anatomy can be expressed by gene \{gene name\}?}
        \item \textit{What anatomy can be upregulated by gene \{gene name\}?}
        \item \textit{How many resemble compounds do \{compound name\} have?}
        \item \textit{How many resemble disease do \{disease name\} have?}
        \item \textit{How many compounds can be used to treat \{disease name\}?}
    \end{itemize}
    \item \textbf{Medium}
        \begin{itemize}
            \item \textit{What compound can treat both \{disease name1\} and \{disease name2\}?}
            \item \textit{What disease located in \{anatomy name\} can \{compound name\} palliate?}
            \item \textit{What disease located in \{anatomy name\} can \{compound name\} treat?}
            \item \textit{What disease is downregulated by \{gene name\} and located in \{anatomy name\}?}
            \item \textit{What disease is associated by \{gene name\} and located in \{anatomy name\}?}
            \item \textit{What disease is upregulated by \{gene name\} and located in \{anatomy name\}?}
            \item \textit{Is there a correlation between \{gene name\} and \{symptom name\}? Please answer True or False}
            \item \textit{Which pharmacologic class includes the most compounds that can palliate the disease with \{symptom name\}?}
            \item \textit{Which pharmacologic class includes the most compounds that can treat the disease with \{symptom name\}?}
            \item \textit{Which cellular component is participated by most genes that are upregulated in disease with \{symptom name\}?}
            \item \textit{Which cellular component is participated by most genes that are associated in disease with \{symptom name\}?}
            \item \textit{Which cellular component is participated by most genes that are downregulated in disease with \{symptom name\}?}
            \item \textit{Which pathway is participated by most genes that are upregulated in disease with \{symptom name\}?}
            \item \textit{Which pathway is participated by most genes that are associated in disease with \{symptom name\}?}
            \item \textit{Which pathway is participated by most genes that are downregulated in disease with \{symptom name\}?}
            \item \textit{How many genes participate the exact same biological processes with \{gene name\}?}
            \item \textit{How many diseases present the exact same symptoms with \{disease name\}?}
        \end{itemize}
\end{itemize}

\paragraph{Legal Graph.}
\begin{itemize}
    \item \textbf{Easy}
    \begin{itemize}
        \item \textit{what is the start date of court \{court name\}?}
        \item \textit{what is the end date of court \{court name\}?}
        \item \textit{what is the citation string of court \{court name\}?}
        \item \textit{which court is handling the case listed under the PACER docket number \{pacer id\}?}
        \item \textit{Who are the attorneys for the case corresponding to this opinion cluster: \{opinion cluster text\}?}
        \item \textit{How many dockets have been processed in court \{court name\}?}
        \item \textit{How many opinions are citing this opinion: \{opinion text\}?}
    \end{itemize}
    \item \textbf{Medium}
        \begin{itemize}
            \item \textit{Which members of the judiciary are responsible for the group of rulings that includes the following opinion: \{opinion text\}}
            \item \textit{What docket includes this opinion: \{opinion plain text\}? Please answer with the pacer case ID.}
            \item \textit{Which court is this opinion cluster syllabus published: \{opinion cluster text\}?}
            \item \textit{How many times has the case \{case name\} been judged in different courts?}
            \item \textit{How many opinions are contained in the opinion clusters about \{case name\}?}
            \item \textit{How many opinions are contained in the opinion cluster with syllabus: \{opinion cluster text\}?}
            \item \textit{How many opinions are contained in the opinion cluster with opinion \{opinion text\}?}
            \item \textit{Which court is this opinion (\{opinion text\}) published?}
            \item \textit{What is the preferred court to cite of judges in court \{source court name\}?}
        \end{itemize}
    \item \textbf{Hard}
        \begin{itemize}
            \item \textit{Is the given sentence supported by the given case? Sentence: {text}, case: \{case name\}.}
            \item \textit{Find a case which can support this sentence: \{text\}.}
        \end{itemize}
\end{itemize}

\section{Question Template Paraphrase Prompt}\label{apx:sec:paraphrase-prompt}

\begin{mybox}{Paraphrase Prompt}
Paraphrase the given template in four different ways. Keep the name in `{}' unchanged, don't use ' in question, and use the same format (`question string', `answer string'):
\end{mybox}

\section{Programmatic Automatic Answer Generation Examples}\label{apx:sec:answer}

\begin{lstlisting}[language=Python]
# Define graph walking functions
def one_hop(graph, center_node_type, center_node_save_key, neighbor_node_type, neighbor_node_save_key, edge_type, k):
    generated_data = []
    cnt = 0
    center_ids = list(graph[center_node_type].keys())
    random.shuffle(center_ids)
    for center_id in center_ids:
        center_name = graph[center_node_type][center_id]['features']['name']
        if edge_type not in graph[center_node_type][center_id]['neighbors']:
            continue
        neighbor_ids = graph[center_node_type][center_id]['neighbors'][edge_type]
        neighbor_names = [graph[neighbor_node_type][neighbor_id]['features']['name'] for neighbor_id in neighbor_ids]
        if len(neighbor_names) > 5:
            continue
        generated_data.append({center_node_save_key:center_name, neighbor_node_save_key: ', '.join(neighbor_names)})
        cnt += 1
        if cnt == k:
            break
    return generated_data

# Generate examples
random.seed(2023)
question = "what are the side effects of compound {compound_name}?"
answer = "{side_effects}"
generated_data = one_hop(graph, 'Compound_nodes', 'compound_name', 'Side_Effect_nodes', 'side_effects', 'Compound-causes-Side Effect', k)
assert len(generated_data) == k
all_generated_data[(question, answer)] = generated_data
\end{lstlisting}

\section{Prompts in \Ours}\label{apx:sec:ours-prompts}

The prompt to instruct LLMs for \Ours contains three parts: graph description, interaction function description, and demonstrations.
The final prompt is shown below, where ``graph definition'', ``interaction function descriptions'' and ``examples'' correspond to the three parts respectively: 

\begin{mybox}{\Ours prompt}

Solve a question answering task with interleaving Thought, Interaction with Graph, Feedback from Graph steps. 

In Thought step, you can think about what further information is needed, and In Interaction step, you can get feedback from graphs with four functions: 

\{interaction function descriptions\}

You may take as many steps as necessary.

Here are some examples:

\{examples\}

(END OF EXAMPLES)

Definition of the graph: \{graph definition\}

Question: \{question\} 

Please answer by providing node main feature (e.g., names) rather than node IDs.
\end{mybox}

 

\subsection{Graph Description Prompts}

\begin{mybox}{MAG graph descriptions}

There are three types of nodes in the graph: paper, author and venue. Paper nodes have features: title, abstract, year and label. Author nodes have features: name. Venue nodes have features: name. Paper nodes are linked to author nodes, venue nodes, reference nodes and cited by nodes. Author nodes are linked to paper nodes. Venue nodes are linked to paper nodes.
\end{mybox}

\begin{mybox}{DBLP graph descriptions}

There are three types of nodes in the graph: paper, author and venue. Paper nodes have features: title, abstract, keywords, lang, and year. Author nodes have features: name and organization. Venue nodes have features: name. Paper nodes are linked to their author nodes, venue nodes, reference nodes (the papers this paper cite) and cited by nodes (other papers which cite this paper). Author nodes are linked to their paper nodes. Venue nodes are linked to their paper nodes.\end{mybox}

\begin{mybox}{E-commerce graph descriptions}

There are two types of nodes in the graph: item and brand. Item nodes have features: title, description, price, img, category. Brand nodes have features: name. Item nodes are linked to their brand nodes, also viewed item nodes, buy after viewing item nodes, also bought item nodes, bought together item nodes. Brand nodes are linked to their item nodes.
\end{mybox}

\begin{mybox}{Literature graph descriptions}

There are four types of nodes in the graph: book, author, publisher, and series. Book nodes have features: country code, language code, is ebook, title, description, format, num pages, publication year, url, popular shelves, and genres. Author nodes have features: name. Publisher nodes have features: name. Series nodes have features: title and description. Book nodes are linked to their author nodes, publisher nodes, series nodes and similar books nodes. Author nodes are linked to their book nodes. Publisher nodes are linked to their book nodes. Series nodes are linked to their book nodes.
\end{mybox}

\begin{mybox}{Healthcare graph descriptions}

There are eleven types of nodes in the graph: Anatomy, Biological Process, Cellular Component, Compound, Disease, Gene, Molecular Function, Pathway, Pharmacologic Class, Side Effect, Symptom. Each node has name feature. There are these types of edges: Anatomy-downregulates-Gene, Anatomy-expresses-Gene, Anatomy-upregulates-Gene, Compound-binds-Gene, Compound-causes-Side Effect, Compound-downregulates-Gene, Compound-palliates-Disease, Compound-resembles-Compound, Compound-treats-Disease, Compound-upregulates-Gene, Disease-associates-Gene, Disease-downregulates-Gene, Disease-localizes-Anatomy, Disease-presents-Symptom, Disease-resembles-Disease, Disease-upregulates-Gene, Gene-covaries-Gene, Gene-interacts-Gene, Gene-participates-Biological Process, Gene-participates-Cellular Component, Gene-participates-Molecular Function, Gene-participates-Pathway, Gene-regulates-Gene, Pharmacologic Class-includes-Compound.
\end{mybox}

\begin{mybox}{Legal graph descriptions}

There are four types of nodes in the graph: opinion, opinion cluster, docket, and court. Opinion nodes have features: plain text. Opinion cluster nodes have features: syllabus, judges, case name, attorneys. Docket nodes have features: pacer case id, case name. Court nodes have features: full name, start date, end date, citation string. Opinion nodes are linked to their reference nodes and cited by nodes, as well as their opinion cluster nodes. Opinion cluster nodes are linked to opinion nodes and docket nodes. Docket nodes are linked to opinion cluster nodes and court nodes. Court nodes are linked to docket nodes.
\end{mybox}

\subsection{Interaction Function Description Prompts}

\begin{mybox}{Interaction function descriptions}

(1) RetrieveNode[keyword], which retrieves the related node from the graph according to the corresponding query.

(2) NodeFeature[Node, feature], which returns the detailed attribute information of Node regarding the given "feature" key.

(3) NodeDegree[Node, neighbor type], which calculates the number of "neighbor type" neighbors of the node Node in the graph.

(4) NeighbourCheck[Node, neighbor type], which lists the "neighbor type" neighbours of the node Node in the graph and returns them.
\end{mybox}

\subsection{Demonstrations}

In \Ours, we provide three demonstrations to teach LLMs how to utilize the four interaction functions.
The demonstrations for academic domain graphs are shown in Figure \ref{apx:fig:demonstration}. More detailed information about demonstrations for other domain graphs can be found at \url{https://github.com/PeterGriffinJin/Graph-CoT/blob/main/Graph-CoT/code/graph_fewshots.py}.

\begin{figure*}[ht!]

\begin{mybox}{Academic graph demonstrations}

Definition of the graph: \{academic graph definition\}

Question: When was the paper Strongly Interacting Higgs Sector in the Minimal Standard Model published?

Reasoning 1: The question is asking some basic information of a node (Strongly Interacting Higgs Sector in the Minimal Standard Model). We need to find the node in the graph.

Interaction 1: RetrieveNode[Strongly Interacting Higgs Sector in the Minimal Standard Model]

Execution 1: The ID of this node is 3101448248.

Reasoning 2: The question is asking the published date of a paper, we need to check the node feature (year) from the graph.

Interaction 2: NodeFeature[3101448248, year]

Execution 2: 1993

Reasoning 3: The published date of the paper is 1993.

Interaction 3: Finish[1993]

\vspace{10pt}
Definition of the graph: \{academic graph definition\}

Question: How many authors do the paper Mass Accretion Rates in Self-Regulated Disks of T Tauri Stars have?

Reasoning 1: The question is asking information of a node (Mass Accretion Rates in Self-Regulated Disks of T Tauri Stars). We need to find the node in the graph.

Interaction 1: RetrieveNode[Mass Accretion Rates in Self-Regulated Disks of T Tauri Stars]

Execution 1: The ID of this node is 2090642949.

Reasoning 2: The question is asking the number of authors of a paper, we need to calculate the node's author neighbor degree from the graph.

Interaction 2: NodeDegree[2090642949, author]

Execution 2: 2

Reasoning 3: The number of the authors is 2

Interaction 3: Finish[2]

\vspace{10pt}
Definition of the graph: \{academic graph definition\}

Question: What was the publish venue of the paper Mass Accretion Rates in Self-Regulated Disks of T Tauri Stars?

Reasoning 1: The question is asking information of a node (Mass Accretion Rates in Self-Regulated Disks of T Tauri Stars). We need to find the node in the graph.

Interaction 1: RetrieveNode[Mass Accretion Rates in Self-Regulated Disks of T Tauri Stars]

Execution 1: The ID of this node is 2090642949.

Reasoning 2: The question is asking the published venue of a paper, we need to check the node's venue neighbor from the graph.

Interaction 2: NeighbourCheck[2090642949, venue]

Execution 2: ['1980519', '1053242']

Reasoning 3: The ID of the published venue are 1980519 and 1053242. We need to get their names.

Interaction 3: NodeFeature[1980519, name], NodeFeature[1053242, name]

Execution 3: the astrophysical journal, the atmosphere journal

Reasoning 4: The name of the published venues are the astrophysical journal and the atmosphere journal

Interaction 4: Finish[the astrophysical journal, the atmosphere journal]
\end{mybox}
\caption{Demonstrations for the academic domain graphs}
\label{apx:fig:demonstration}
\end{figure*}

\end{document}